\documentclass{article}
\usepackage{arxiv}
\usepackage[utf8]{inputenc} 
\usepackage[T1]{fontenc}    
\usepackage{hyperref}       
\usepackage{url}            
\usepackage{booktabs}       
\usepackage{graphicx}
\usepackage{doi}

\usepackage{algorithmic}
\usepackage[ruled,linesnumbered]{algorithm2e}
\usepackage{latexsym}
\usepackage{amssymb}
\usepackage{amsmath}
\usepackage{amsthm}
\usepackage{booktabs}
\usepackage{subcaption}
\usepackage{graphicx}
\usepackage{multirow}
\usepackage{color}
\usepackage{tikz}
\usepackage{comment}

\DeclareMathOperator{\sign}{sign}

\title{Hybrid Real- and Complex-valued Neural Network Architecture}

\author{Alex Young~\thanks{Corresponding authors.} \\
	NXP Semiconductors\\
	High Tech Campus\\
	5656 AE Eindhoven, The Netherlands\\
	\texttt{alex.young@nxp.com} \\
	\And
	Luan Vinícius Fiorio~$ ^*$ \\
	Department of Electrical Engineering\\
	Eindhoven University of Technology\\
	5600 MB Eindhoven, The Netherlands \\
	\texttt{l.v.fiorio@tue.nl} \\
	\And
	Bo Yang \\
	Department of Electrical Engineering\\
	Eindhoven University of Technology\\
	5600 MB Eindhoven, The Netherlands \\
	\And
    Boris Karanov\\ 
    Communications Engineering Lab\\ 
    Karlsruhe Institute of Technology\\
    D-76187 Karlsruhe, Germany
	\And
    Wim van Houtum\\
	NXP Semiconductors\\
	High Tech Campus\\
	5656 AE Eindhoven, The Netherlands\\
    \And
	Ronald M. Aarts\\
	Department of Electrical Engineering\\
	Eindhoven University of Technology\\
	5600 MB Eindhoven, The Netherlands \\     
}

\renewcommand{\shorttitle}{\textit{arXiv} Template}

\hypersetup{
pdftitle={A template for the arxiv style},
pdfsubject={q-bio.NC, q-bio.QM},
pdfauthor={David S.~Hippocampus, Elias D.~Striatum},
pdfkeywords={First keyword, Second keyword, More},
}

\begin{document}
\maketitle

\begin{abstract}
We propose a \emph{hybrid} real- and complex-valued \emph{neural network} (HNN) architecture, designed to combine the computational efficiency of real-valued processing with the ability to effectively handle complex-valued data. We illustrate the limitations of using real-valued neural networks (RVNNs) for inherently complex-valued problems by showing how it learnt to perform complex-valued convolution, but with notable inefficiencies stemming from its real-valued constraints. To create the HNN, we propose to use building blocks containing both real- and complex-valued paths, where information between domains is exchanged through domain conversion functions. We also introduce novel complex-valued activation functions, with higher generalisation and parameterisation efficiency. HNN-specific architecture search techniques are described to navigate the larger solution space. Experiments with the AudioMNIST dataset demonstrate that the HNN reduces cross-entropy loss and consumes less parameters compared to an RVNN for all considered cases. Such results highlight the potential for the use of partially complex-valued processing in neural networks and applications for HNNs in many signal processing domains.
\end{abstract}


\section{Introduction}


The necessity for complex-valued data processing, in many fields of science where deep learning is applied, usually requires data to be converted into a real-valued representation due to limitations neural networks (NNs) have with complex-valued parameters.

For example, in audio processing, discarding the phase component simplifies data handling but reduces audio intelligibility and limits synthesis quality, as phase information is crucial for accurate sound reconstruction \cite{sarroff2018complex}. Even when phase is retained, processing real and imaginary parts separately can cause phase distortion, degrading signal quality \cite{lee2022complex}. 

Deep complex-valued neural networks (CVNNs) have been applied recently to solving traditionally complex-valued problems, e.g., radar \cite{fuchs2021complex} and audio processing \cite{welker2022interspeech}, and even real-valued problems like image processing \cite{quan2021image}, presenting performance gains over RVNNs in most cases. Moreover, complex-valued processing in NNs has shown richer representation capacity with better generalisation characteristics, additionally allowing for faster learning and noise-robust memory mechanisms \cite{trabelsi2018deep}. Automatic differentiation and other overheads within CVNNs typically incur an increased training time compared to using standard functions within an RVNN \cite{lee2022complex}.


The combination of real- and complex-valued processing, in neural network architectures, was previously considered but scarcely explored. Averaging the results of an RVNN and its complex-valued counterpart has been investigated \cite{popa2018deep} -- possibly the simplest approach towards the idea of a hybrid real- and complex-valued architecture. On the other hand, \cite{du2023ahybrid} created a network with complex-valued layers followed by real-valued layers. In both cases, the hybrid version displayed better results than real- or complex-valued counterparts. A two-stage framework has also been explored \cite{ju2023tea}, where an RVNN is followed by a complex-valued processing stage. Similarly, \cite{lv2021dccrn} applied additional real-valued processing to a complex-valued neural network. However, the interchange of information between both domains was not explored and no systematic approach was provided in the design of a hybrid architecture. For this reason, one of the main contributions of our work is to propose a comprehensive framework for designing HNNs.

We present a novel and flexible neural network architecture that systematically combines the strengths of both real-valued and complex-valued processing. By integrating pathways for both, this approach allows us to select optimal activation functions and processing methods in each domain to minimise loss and parameter storage. We also devised a neural architecture search (NAS) procedure such that redundancies between paths can be reduced together with a decrease in the number of parameters and operations. Note that, when referring to parameters, we count two parameters per complex weight or bias and one per real weight or bias.

\section{Architecture}
\label{sec:architecture}

The HNN architecture was inspired by observations of complex-valued processes in RVNNs when dealing with complex-valued problems. A complex-valued input was connected to an RVNN by splitting and interleaving the real and imaginary parts. By plotting the weights of a fully connected layer in this network and re-ordering the outputs by similarity, we noticed a weight value behaviour that corresponded, in part, to a complex-valued convolution. Next, we describe a straightforward example problem where this behaviour can be observed.

\subsection{RVNN experiment}
\label{sec:RVNN_experiment}

To estimate the parameters of a complex-valued sinusoidal signal with noise, consider the model $y = a e^{i(2\pi nm/N + p)}+v(n)$ with ranges for amplitude $a \in \{0.1,1.0\}$, frequency in cycles $m \in \{5.0, 12.0\}$, phase in radians $p \in \{-\pi, \pi\}$, complex-valued noise $v(n)$ with magnitude $\in \{0,0.01\}$ and $n=0...(N-1)$. With $N=512$, the signal $y$ is multiplied by a Hanning window and converted to the frequency domain using an $N$ point discrete Fourier transform, a training set of 500k samples is generated using the defined parameter ranges. Only the first 18 frequency bins are used, splitting and interleaving the real and imaginary parts to form 36 real-valued predictors. The objective is to estimate [$m$, $a \sin{(p)}$, $a \cos{(p)}$].

\begin{figure}
    \centering
    \includegraphics[width=0.4\linewidth]{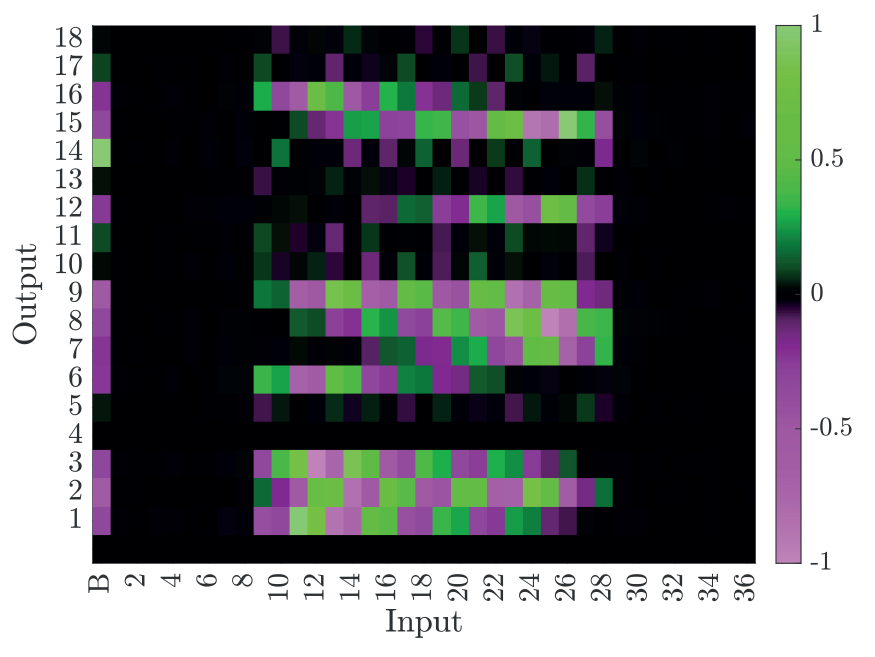}
    \caption{Non-ordered weights for the first convolutional layer.}
    \label{fig:nonordered_weights}
\end{figure}

\begin{figure}[!t]
    \centering
    \includegraphics[width=0.7\linewidth]{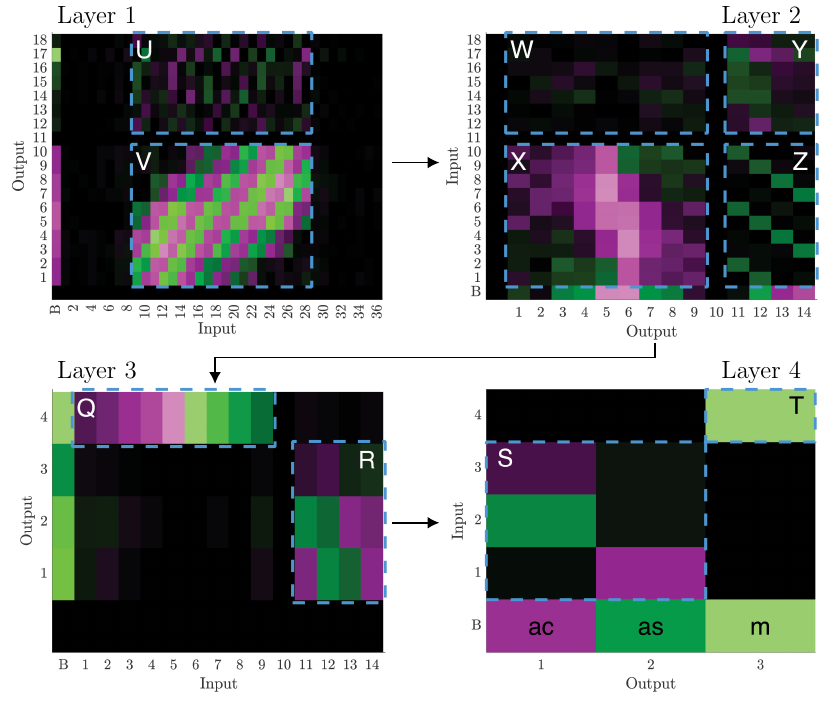}
    \caption{Plot of reordered and normalized weights for each layer's convolutions for the RVNN experiment. The arrows indicate how the output of each layer is followed by the input of another. Distinct regions in the weight patterns are highlighted with dashed blue lines, presenting a letter as its label. Refer to the colour bar in Figure~\ref{fig:nonordered_weights}.}
    \label{fig:layers_regions}
\end{figure}

Moreover, we defined an NN model consisting of four fully connected layers, where the first three are followed by exponential linear unit (ELU) activation functions. The output sizes of the four layers are 18, 14, 4 and 3, respectively. The NN model was trained multiple times to verify that it exhibits similar weight behaviour. Figure~\ref{fig:nonordered_weights} shows the weights of the first fully connected layer mapped from input to output, obtained after 2500 epochs of training and normalised from -1 to 1, where the first column is the bias. No obvious conclusions can be drawn from such a plot, except from the insight that some outputs follow a similar pattern. By reordering the outputs, placing similar weight rows together, as per Figure~\ref{fig:layers_regions} (Layer 1 for a direct comparison to Figure~\ref{fig:nonordered_weights}, we can observe a very strong pattern in the weight values, mainly from outputs 1 to 10. Extending the reordering to all convolutions, as shown in Figure~\ref{fig:layers_regions}, we can identify separated regions within each convolution.

\begin{figure}[!t]
    \centering
    \includegraphics[width=1.0\linewidth]{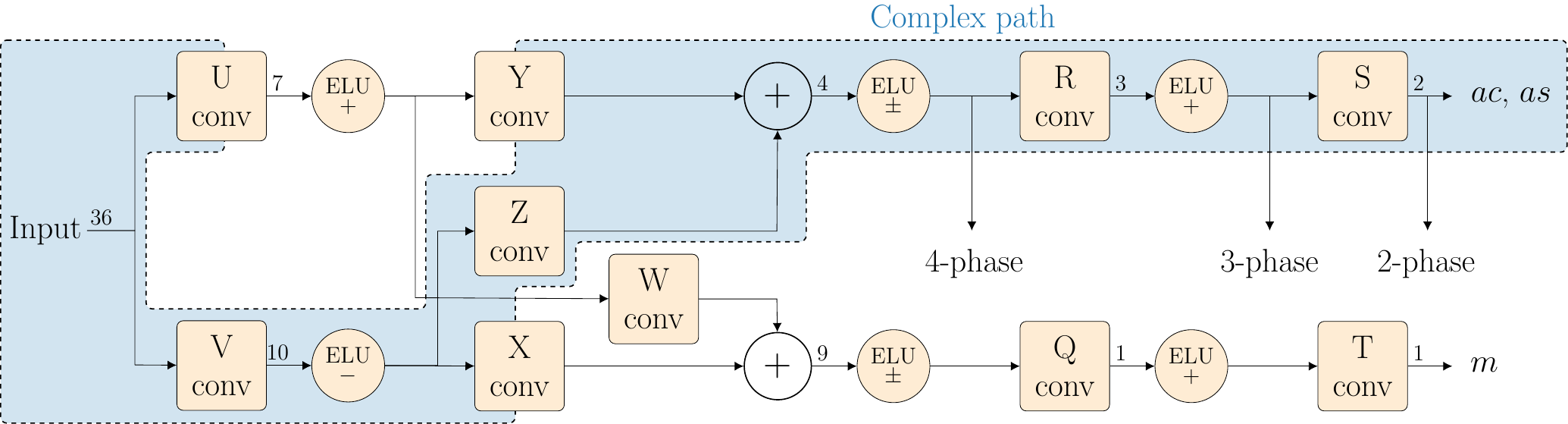}
    \caption{Detailed schematic of the trained and decoded RVNN used for the experiment, given weight regions defined in Figure~\ref{fig:layers_regions}. The complex path shows the activations that represent the path where complex-valued data is conveyed in various forms, learned by the RVNN. When a number is indicated after the input, a layer, or a summation, it indicates size.}
    \label{fig:experiment_schematic}
\end{figure}

\begin{figure}[!t]
\centering
    \begin{subfigure}{.32\textwidth}
        \centering
        \includegraphics[width=1\textwidth]{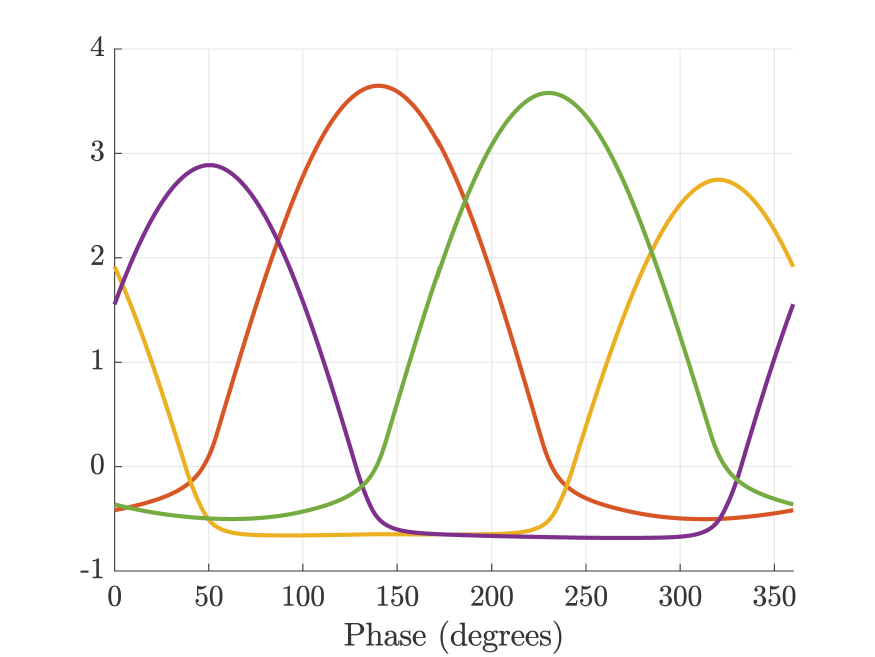}  
        \caption{4-phase}
        \label{fig:4phase}
    \end{subfigure}
    \begin{subfigure}{.32\textwidth}
        \centering
        \includegraphics[width=1\textwidth]{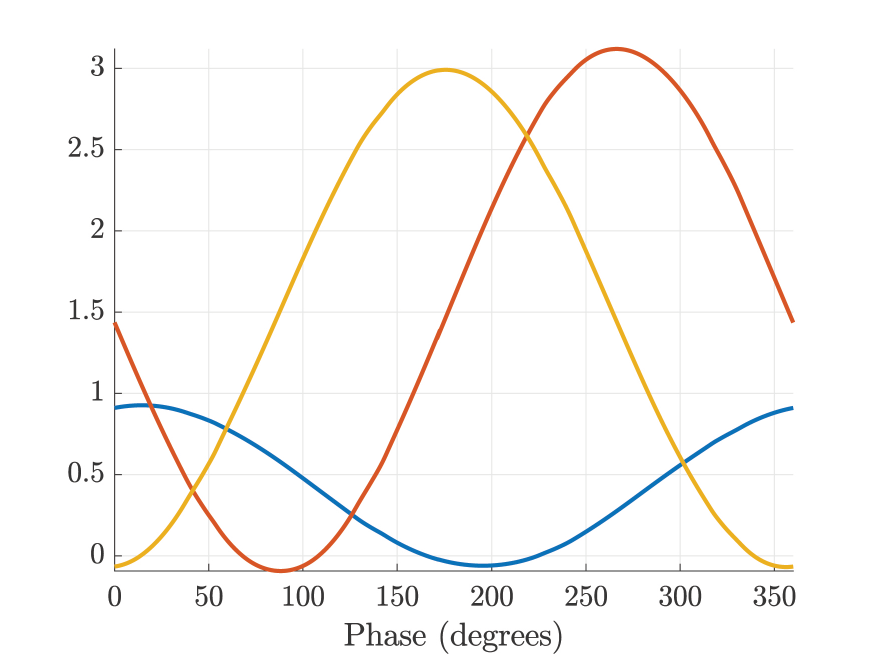}
        \caption{3-phase}
        \label{fig:3phase}
    \end{subfigure}
    \begin{subfigure}{.32\textwidth}
        \centering
        \includegraphics[width=1\textwidth]{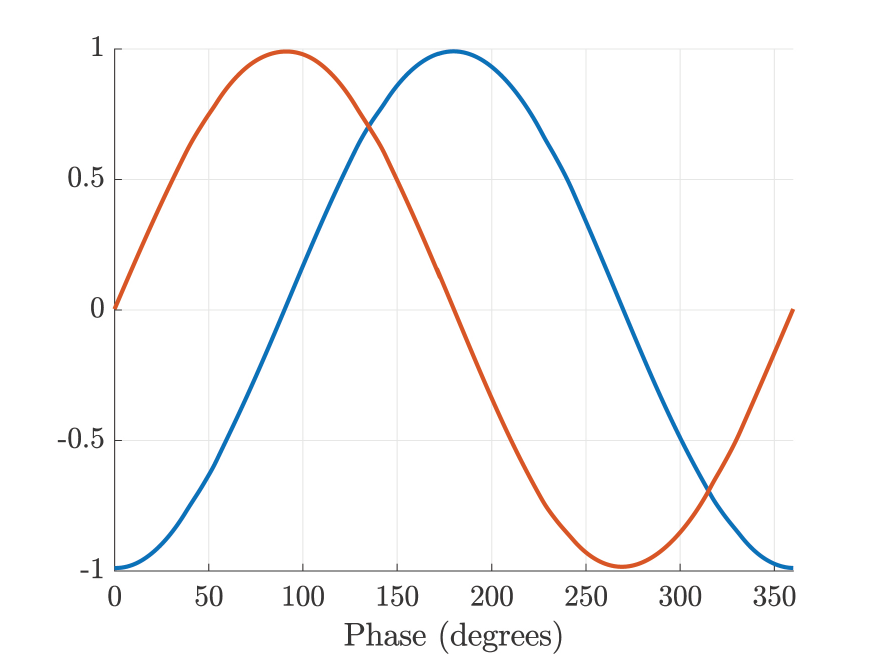}
        \caption{2-phase}
        \label{fig:2phase}
    \end{subfigure}
\caption{Multi-phase real-valued functions observed in the 4-, 3- and 2-phase indication in Figure~\ref{fig:experiment_schematic}.}
\label{fig:phases}
\end{figure}

To better understand the flow of information in the RVNN, we look at the normalised and re-ordered weights of all convolutional layers, as shown in Figure~\ref{fig:layers_regions}. Figure~\ref{fig:experiment_schematic} shows the identified weight regions Q to Z as small convolutions shown as square shapes and the associated information pathways, we highlight the `complex path' to identify the signals that ultimately form the complex-valued output pair `ac' and `as'. The downward arrows take various points along this path to show the different variations that the signal can take, with the complex-valued output showing a 2-phase representation, 3-phase and 4-phase signals are also shown in Figure~\ref{fig:phases}. The 3- and 4-phase representations were used as an inspiration for the domain conversion functions mentioned later.
Activation functions, shown as circular shapes, indicate the type of function and the general bias level on it's inputs, obtained by observing the sign of the bias from the preceding convolution. In this case all activations have been selected to be ELU. The bias sign can be inferred from the weight maps with the label `B', for example in Figure~\ref{fig:layers_regions}, the first column adjacent to the V region is purple indicating a negative bias so the following ELU activation function will have a negative bias moving a zero-mean input signal into the non-linear region. 

\begin{figure}[!t]
    \centering
    \includegraphics[width=0.4\linewidth]{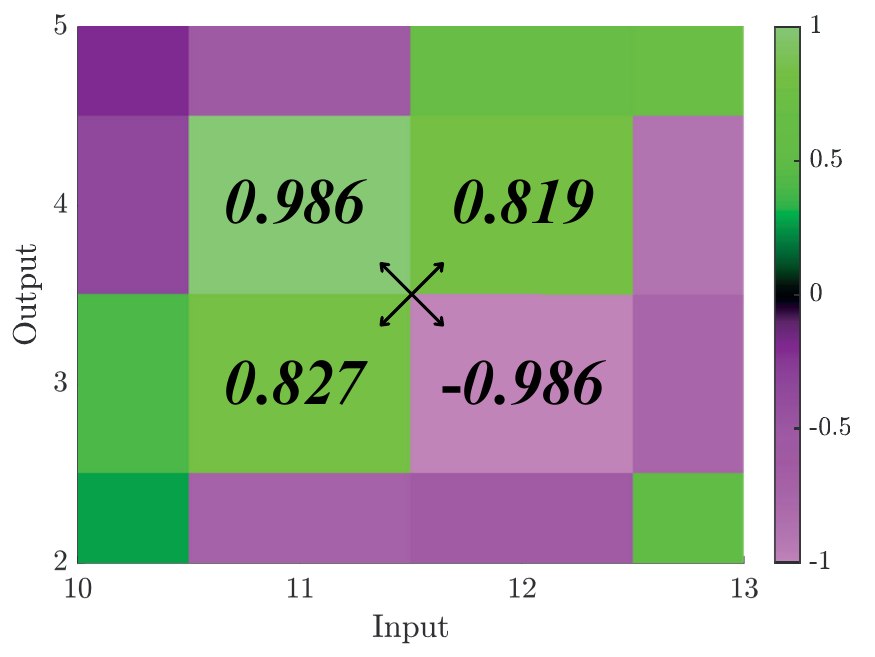}
    \caption{Detail view of reordered weights of the first convolutional layers in the RVNN network considered for the experiment.}
    \label{fig:weights_zoom}
\end{figure}

To better show how the reordered weights represent a complex convolution, we zoom into a small section of Figure~\ref{fig:layers_regions}, for the first layer, obtaining Figure~\ref{fig:weights_zoom}, which is analysed next. Let $\boldsymbol{in} = [in_{11} \ \ in_{12}]^{T}$, where the superscript $^T$ indicates the transpose operation, represent the input signal vector for inputs 11 (real) and 12 (imaginary). The weights with values depicted in Figure~\ref{fig:weights_zoom} map inputs 11 and 12 to an output vector $\boldsymbol{out} = [out_3 \ \ out_4]^{T}$ with a weight matrix 
\begin{equation}
    \boldsymbol{W} = 
    \begin{bmatrix}
        0.827 & -0.986\\
        0.986 & 0.819\\
    \end{bmatrix}.
\end{equation}
We can define the output as
\begin{equation}
    \begin{bmatrix}
        out_3 \\
        out_4
    \end{bmatrix}
    =
    \boldsymbol{W} \times \boldsymbol{in}
    =
    \begin{bmatrix}
        0.827 & -0.986\\
        0.986 & 0.819\\
    \end{bmatrix}
    \times
    \begin{bmatrix}
        in_{11} \\
        in_{12}
    \end{bmatrix}.
\label{eq:network_weights}
\end{equation}

On the other hand, let us define $x = x_r + i x_i$ and $w = w_r + i w_i$ as complex numbers. The complex-valued convolution between $x$ and $w$ results in a complex number $o = o_r + io_i$ which can be obtained as $o = w \ast x$, where
\begin{equation}
    \begin{bmatrix}
        o_r \\
        o_i
    \end{bmatrix}    
    = 
    \begin{bmatrix}
        w_r & -w_i\\
        w_i & w_r
    \end{bmatrix}
    \times
    \begin{bmatrix}
        x_r \\
        x_i
    \end{bmatrix}.
\label{eq:complex_convolution}
\end{equation}
With the similarity of \eqref{eq:network_weights} and \eqref{eq:complex_convolution}, $out_3$ and $out_4$ are equivalent to $o_r$ and $o_i$, as well as $in_{11}$ and $in_{12}$ to $x_r$ and $x_i$. We observe that $\boldsymbol{W}$ contains a good approximation of the complex conjugates of a complex number. In this sense, we can interpret \eqref{eq:network_weights} as an approximate representative part of a complex-valued convolution.

This shows that the neural network, in part, is inherently learning to approximate a complex-valued convolution using real-valued weights. Maybe it is inefficient for the real-valued network to recreate complex-valued convolutions as four real-valued weights are required for each complex-valued multiplication, whereas in an efficient complex-valued multiplication, only two weights are required with the associated reduction in memory space and access. Additionally, outputs 11 to 18 in Figure~\ref{fig:layers_regions}, Layer 1, do not show such patterns consistent with complex-valued convolution, simply representing a real-valued convolution. 

For this use case, we see the necessity of adding complex-valued processing to the neural network, while still allowing for real-valued paths.

Using the trained network for inference with generated datasets that kept the amplitude and frequency model parameters constant but swept the phase parameter between $-\pi$ and $+\pi$, we were able to observe how a rotating phasor at the output was internally represented. We saw that 4- and 3-phase activations with a positive value were present as the information was transformed into the phasor outputs. This presented the first insights into how a complex value could be mapped into a multi-dimensional positive real space and vice-versa. Such a mapping is taken into account as a possible conversion function and outlined in Section~\ref{sec:domain_conversion}.

Allowing the network to have paths of complex-valued processing while still maintaining real-valued branches would be a natural solution to facilitate complex-valued processing and avoid the disadvantages that a pure CVNN would bring. 
Next, we present a hybrid real- and complex-valued building block for convolutional/fully connected neural networks, which we use to construct the HNN model.

\subsection{Building block}
\label{ssec:building-block}

The HNN is built upon functional blocks with real- and complex-valued inputs and outputs. For generalisation, separate pathways allow all inputs to connect to all outputs using real-to-complex and complex-to-real domain conversion functions. Such functions facilitate the information flow between the two domains and are explored in detail in Section~\ref{sec:domain_conversion}. Each path, real- or complex-valued, can have a convolution with an activation function and additional (optional) functions such as pooling, normalisation and dropout. Complex-valued paths use complex-valued functions and real-valued paths use real-valued functions.

\begin{figure}
    \centering
    \includegraphics[width=0.5\textwidth]{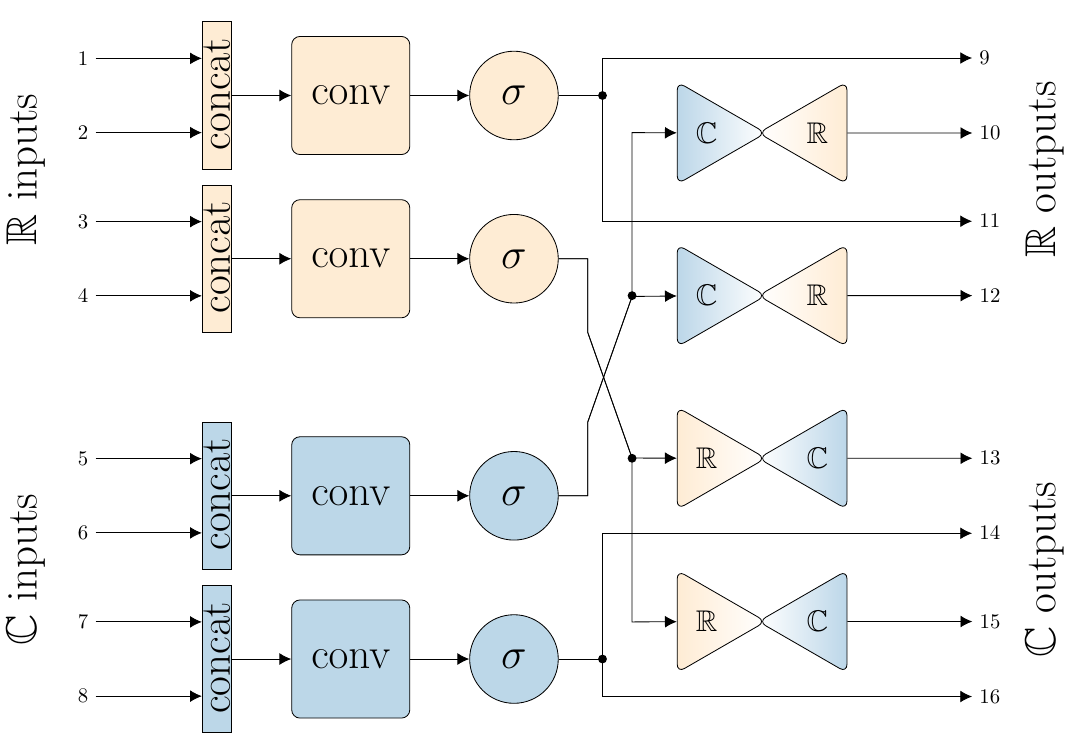}
    \caption{HNN functional block.}
    \label{fig:hybrid_block}
\end{figure}

The diagram of a functional block can be seen in Figure~\ref{fig:hybrid_block}, where inputs and outputs are numbered for further simplification. The ``concat'' rectangles represent concatenations, while ``conv'' stands for convolution, $\sigma$ contains the activation function and other optional functions and the hourglass-shaped blocks ($\mathbb{C} \rightarrow \mathbb{R}$ and $\mathbb{R} \rightarrow \mathbb{C}$) are the domain conversion functions. In the figure, yellow-coloured blocks stand for real-valued processing and blue-coloured blocks stand for complex-valued processing. Importantly, the use of multiple pathways allows for generalisation of the architecture, which can be optimised as described in Section~\ref{sec:architecture_search}. Multiple building blocks can be combined in series and/or parallel. In this work, we specifically try to build a network by connecting such blocks in series.

If the system input data is only real- or only complex-valued, then the unused ports can be either removed -- with the associated routing in the next block -- or domain conversions can be performed from the used input to the other domain input. Similarly, if the system outputs are not required to be both real- and complex-valued, the unused output can be removed together with all dependent routing.


Furthermore, the initial step in designing a HNN is to form a base architecture composed of multiple building blocks. Such an architecture can be inherited from a real-valued NN model, substituting each convolutional/fully connected layer for the functional block in Figure~\ref{fig:hybrid_block}, or defined from scratch with multiple interconnected blocks.

The architecture can be tailored for the problem at hand, in the case that the type of input and output are available as prior knowledge. For example, if the input is the STFT of a signal, the complex-valued inputs 5 and 7 or 6 and 8 from Figure~\ref{fig:hybrid_block} could be used, including domain conversion functions that would transform the complex-valued input into a real-valued input, applied to inputs 1 and 3 or 2 and 4 from the functional block. Propagating forward through such connections in the HNN model determines which functions are used. Similarly, the same can be done for the outputs. Considering, e.g., a real-valued output, a backward dependency check can be done to remove unnecessary connections. 

By defining a base architecture and checking dependencies, we arrive at an initial model which will be subject to a systematic architecture search procedure. This step is needed for reducing the total number of parameters and required processing that the initial hybrid model presents and thus making it feasible for practical implementation. Importantly, the search also automates the choice of activation and domain conversion functions, which is a time-consuming task if performed manually. The next section presents our proposed architecture search scheme for HNNs.

\section{Architecture search}
\label{sec:architecture_search}

\begin{figure}[!t]
    \centering
    \includegraphics[width=0.6\textwidth]{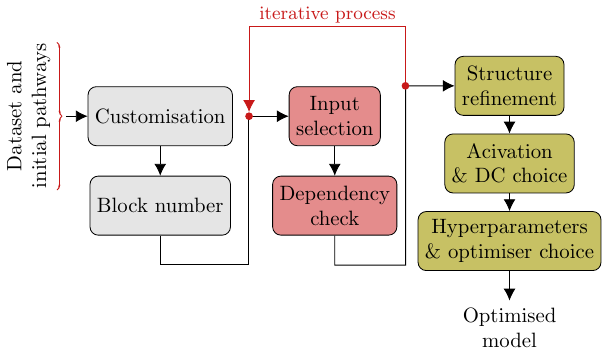}
    \caption{HNN architecture search process.}
    \label{fig:NASworkflow}
\end{figure}

To find an optimal trade-off between performance and complexity, fine-tuned parameters and an efficient network structure are required. Conventionally, this process has relied heavily on time consuming manual design, however, with developments in NAS methods, automated search algorithms are increasingly used. In particular, Optuna, originally developed by \cite{akiba2019optuna} in 2019, is a hyperparameter optimisation framework which returns an optimised neural network configuration by evaluating the performance of different structures and hyperparameters. Thus, we adopted a framework for building HNNs using Optuna. The methodology comprises eight sequential phases aimed at tailoring and refining a HNN to suit a specific task. The sequence of phases is shown in Figure~\ref{fig:NASworkflow} and is described next.

Initially, the input and output pathways of the network are adjusted for the needs of the task, and their dependencies are checked to mitigate redundancy (``Customisation'' phase). The number of blocks (``Block number'' phase) is selected based on performance, where balancing model complexity and computational efficiency is important. Until the block number selection, the optimisation objective regards only validation loss reduction. For the subsequent steps, a limitation in the number of parameters is also included as part of the objective. Additionally, a preliminary learning rate must be established, serving as a starting point.






Whenever the architecture is changed by the NAS, the ``Input Selection'' phase identifies unnecessary paths in all blocks, which have then their dependencies adjusted to reduce redundancy (``Dependency Check'' phase), in an iterative process. After these steps, the architecture is refined in the ``Structure Refinement'' phase, involving the optimisation of blocks within the architecture, including adjustments in convolution channel number and whether other (optional) function layers should be included within each block.

The activation functions and domain conversion functions are chosen in a later phase, named ``Activation \& DC choice''. The activation functions for block $N_b$ are chosen from a candidate list. We also offer the option of ``no activation'' as a selection for functions that precede a domain conversion function, since they can also introduce non-linear transformations. The domain conversion functions are selected from a candidate list, which serve to interchange information between real- and complex-valued pathways.

Finally, the last step includes the optimisation of the learning rate and the selection of a better optimisation algorithm (``Hyperparameter choice'' phase), aimed at enhancing convergence speed and overall efficiency. This methodology facilitates the comprehensive building and optimisation of HNN architectures tailored to a specific task.

\section{Domain conversion}
\label{sec:domain_conversion}
HNNs employ conversion between the real- and complex-domains, this is usually limited to Cartesian or polar conversions. In the RVNN experiment described in Section~\ref{sec:RVNN_experiment}, we also observed that other functions may be useful and thus we widened the function space. For the purposes of building a HNN, the mapping does not need to be invertible and, for generalisation, should be allowed to contract or expand dimensionality with the associated loss or redundancy in information. 

We strive to define diverse forms of conversion from the real- to complex-domain, mostly based on Cartesian and polar functions. Using between one to three real-valued inputs $x_1$, $x_2$ and $x_3$ to produce a complex output $z$, we define seven domain conversion functions.

\begin{subequations}
    \vspace{4mm}
    \text{Single input real- to complex-valued functions (expansion):}
    \begin{equation} \text{Real: }   z=x_1;               \label{eq:R2C_Real}   \end{equation}  
    \begin{equation} \text{Exp: }    z=e^{i\pi x_1};      \label{eq:R2C_Exp}    \end{equation}
    \begin{equation} \text{Sqrt: }   z=\sqrt{x_1};        \label{eq:R2C_Sqrt}   \end{equation}
    \begin{equation} \text{MagExp: } z=|x_1|e^{i\pi x_1}; \label{eq:R2C_MagExp} \end{equation}

    \vspace{4mm}
    \text{Dual input real- to complex-valued functions (no loss of information):}
    \begin{equation} \text{Cartesian: } z=x_1+ix_2;        \label{eq:R2C_Cartesian} \end{equation} 
    \begin{equation} \text{Polar: }     z=x_1e^{i\pi x_2}; \label{eq:R2C_Polar}     \end{equation}

    \vspace{4mm}
    \text{Triple input real- to complex-valued function (contraction):}
    \begin{equation} \text{Rotation: } z=(x_1+ix_2)e^{i\pi x_3}; \label{eq:R2C_Rotation} \end{equation}
\end{subequations}

We define a complimentary set of complex- to real-domain conversion functions producing one to three outputs ($y_1$, $y_2$ and $y_3$) from a complex-valued input $z$, including improved handling of phase information. Using raw phase information can cause problems due to the discontinuity of the argument function as the phase value wraps around $\pm\pi$, a circular loss function may help to mitigate the issue if phase estimation is needed as a target. Alternatively, using the absolute value of the phase eliminates the discontinuity. Nine functions are offered where information is lost (lossy) or where no information is lost (lossless). ($\angle$ represents the argument or angle function).

\begin{subequations}
    \vspace{4mm}
    \text{Lossy functions:}
    \begin{equation} \text{Real: }          y_1=\Re(z);                       \label{eq:C2R_Real}        \end{equation}
    \begin{equation} \text{Mag: }           y_1=|z|;                          \label{eq:C2R_Mag}         \end{equation}
    \begin{equation} \text{SquareMag: }     y_1=|z|^2;                        \label{eq:C2R_SquareMag}   \end{equation}
    \begin{equation} \text{AbsPhase: }      y_1=|\angle(z)|/\pi;              \label{eq:C2R_AbsPhase}    \end{equation}
    \begin{equation} \text{MagAbsPhase: } \{y_1=|z|, \ y_2=|\angle(z)|/\pi\}; \label{eq:C2R_MagAbsPhase} \end{equation}

    \vspace{4mm}
    \text{Lossless functions:}
    \begin{equation} \text{Cartesian: } \{y_1=\Re(z),\ y_2=\Im(z)\};     \label{eq:C2R_Cartesian} \end{equation} 
    \begin{equation} \text{Polar: }     \{y_1=|z|, \ y_2=\angle(z)/\pi\}; \label{eq:C2R_Polar}     \end{equation}

    \vspace{4mm}
    \text{Multi-phase functions (lossless when $N_p \ge 3$):}
    \begin{equation} \text{MultiMagReal: }  y_n = (|z|+\Re(ze^{-2i\pi n/N_p})/2, n=0 ... (N_p-1);    \label{eq:C2R_MultiMagReal} \end{equation}  
    \begin{equation} \text{MultiMagPhase: } y_n = (|z||\angle(ze^{-2i\pi n/N})|)/\pi, n=0...(N_p-1); \label{eq:C2R_MultiMagPhase} \end{equation}
\end{subequations}

The multi-phase functions were inspired by observations in Section~\ref{sec:architecture}, to reproduce the positive biased functions ($y_n \ge 0$), a combination of the magnitude (providing positive offset) and a phase-shifted version of the complex-valued input was used in the first case. Similarly, we define an angle-based conversion function that generates outputs that have a linear phase characteristic.
By equally spacing the phase shifts, any number of outputs can be produced. $N_p$ is the number of output phases, $N_p = 3$ is a fair choice, reducing the preliminary network architecture search space and maintaining a conversion without loss of information.


\section{Complex-valued activation functions}
\label{sec:complex_activation}
Many complex-valued activation functions have been proposed \cite{trabelsi2018deep, bassey2021survey}. Starting from the function $Z_o=z/(|z|+\epsilon)$ \cite{georgiou1992complex}, we first generalised it to include a soft threshold shape when $z$ is real-valued, by including $\alpha$ and $\beta$ parameters, resulting in $Z_o=\alpha z+\beta z/(|z|+\epsilon)$. This also gave us the first indication that similar functions in this class can be used effectively with both real- and complex-valued arguments. The following proposed functions are further generalisations:
\begin{subequations}
    \begin{equation}
        Z_o = \alpha z + \frac{1}{|z|^q + \epsilon}\sum_{n=0}^{p}k_n z^n;
    \label{eq:activation_P}
    \end{equation}
    \begin{equation}
        Z_o = \alpha z + \frac{1}{\sqrt{|z|^q + \epsilon}}\sum_{n=0}^{p}k_n z^n,
    \label{eq:activation_Ps}         
    \end{equation}
\end{subequations}
where $p$ and $q$ are integers, $k_n$ is real- or complex-valued and $\epsilon > 0$.

The activation function \eqref{eq:activation_P} includes approximations or substitutes for the rectified linear unit (ReLU), exponential linear unit (ELU) variants, Tanhshrink and absolute real-valued activation functions. For example, the ELU activation can be approximated with $Z_o=0.373z+z(1.513+0.627z)/(|z|+2.411)$. 
Function \eqref{eq:activation_Ps} includes the real-valued Tanh- and Softplus-shaped functions, respectively. Many familiar real-valued functions can thus be extended to the complex domain via these proposed general forms. 

We can also form two purely complex-valued activation functions to add diversity to the function library:
\begin{equation}
    Z_o = z (1 - |\alpha|+\alpha \sign(v)),
\label{eq:activation_D}   
\end{equation}
which is a complex-valued switching function as another analogue to the real-valued ReLU function; and
\begin{equation}
    Z_o = z \left( 1 - |\alpha| + \frac{\alpha v^q}{|z|^q+\epsilon} \right),
\label{eq:activation_E}          
\end{equation}
which has a smooth transition instead, where $v = \max(0, \Re(z) - k_1 |\Im(z)| - k_0)$.




To reduce the search space during network optimisation, seven basic shapes of complex-valued functions are considered, where five relate to real-valued equivalent functions: ReLU, Tanh, Tanhshrink, SoftPlus and Abs (an uncommon activation function, identified in \cite{sipper2021neural}); and two cases for \eqref{eq:activation_D} and \eqref{eq:activation_E}. The parameters $q$, $k_n$, $\alpha$, $\beta$, $\gamma$ and $\epsilon$, used to achieve the aforementioned shapes, are shown in Table~\ref{tab:activations-parameters} with their proposed names. The parameters of the chosen activations can be tuned in later optimisation processes if desired.

\begin{table}[!t]
    \centering
    \caption{Parameters used to define the shape of the considered complex-valued activation functions.}
    \vspace{1mm}
    \begin{tabular}{c c c c c c c c c}
        \hline
        \textbf{Name} & \textbf{Eq.} & $\boldsymbol{q}$ & $\boldsymbol{\alpha}$ & $\boldsymbol{k_0}$ & $\boldsymbol{k_1}$ & $\boldsymbol{k_2}$ & $\boldsymbol{k_3}$ & $\boldsymbol{\epsilon}$ \\
         \hline      
         cRecip      & \eqref{eq:activation_P}  & 1 & 0     & 0     & 1      & 0     & 0   & 0.01  \\
         cReLU       & \eqref{eq:activation_P}  & 1 & 0.5   & 0     & 0      & 0.5   & 0   & 0.01  \\
         cAbs        & \eqref{eq:activation_P}  & 1 & 0     & 0     & 0      & 1     & 0   & 0.01  \\
         cTanhshrink & \eqref{eq:activation_P}  & 2 & 0     & 0     & 0      & 0     & 1   & 1     \\
         cTanh       & \eqref{eq:activation_Ps} & 2 & 0     & 0     & 1      & 0     & 0   & 1     \\
         cSoftPlus   & \eqref{eq:activation_Ps} & 2 & 0.5   & 2.134 & 0      & 0.5   & 0   & 9.481 \\
         cReImLU     & \eqref{eq:activation_D}  & - & 0.95  & 0.1   & 1      & -     & -   & -     \\
         cRecipMax   & \eqref{eq:activation_E}  & 2 & 0.9   & 0.1   & 0.5    & -     & -   & 0.1   \\
         \hline
    \end{tabular}
    \label{tab:activations-parameters}
\end{table}

\section{Numerical experiments}
\label{sec:numerical-experiment}

To validate the practicality and advantages of the HNN model, we employed the AudioMNIST dataset~\cite{bacher2023audio} as a proof of concept. This dataset comprises 30000 audio samples of spoken digits (0-9) produced by 60 distinct speakers. To evaluate the HNN, we compare it to an optimised RVNN model using a NAS process similar to what is described in Section~\ref{sec:architecture_search}. The primary objective is to evaluate the performance of both the HNN and RVNN models in the classification of spoken digits. 

\begin{figure*}[!t]
    \centering
        \includegraphics[width=1.0\linewidth]{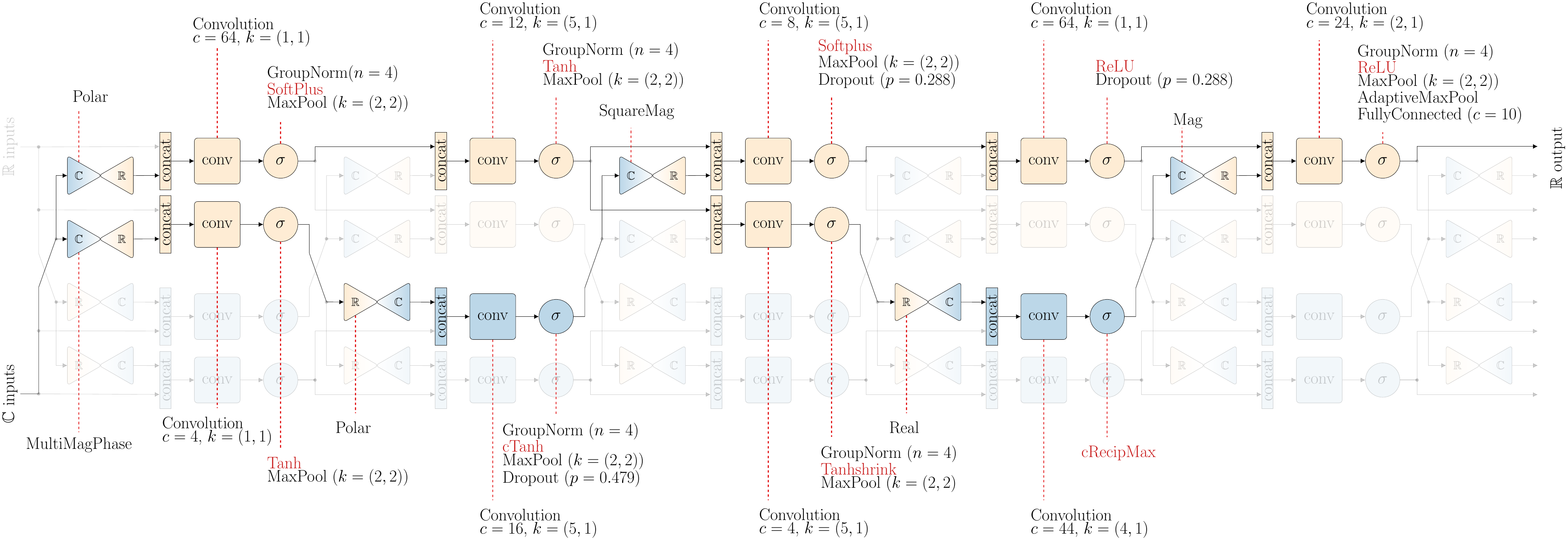}
    \caption{Optimised HNN Architecture for -5 dB SNR noise. $c$ is output channels, $k$ is kernel, $n$ is number of groups and $p$ is dropout rate.}
    \label{fig:Hybrid-valued-nn}
\end{figure*}

\begin{figure}[!t]
    \centering
    \includegraphics[width=1.0\linewidth]{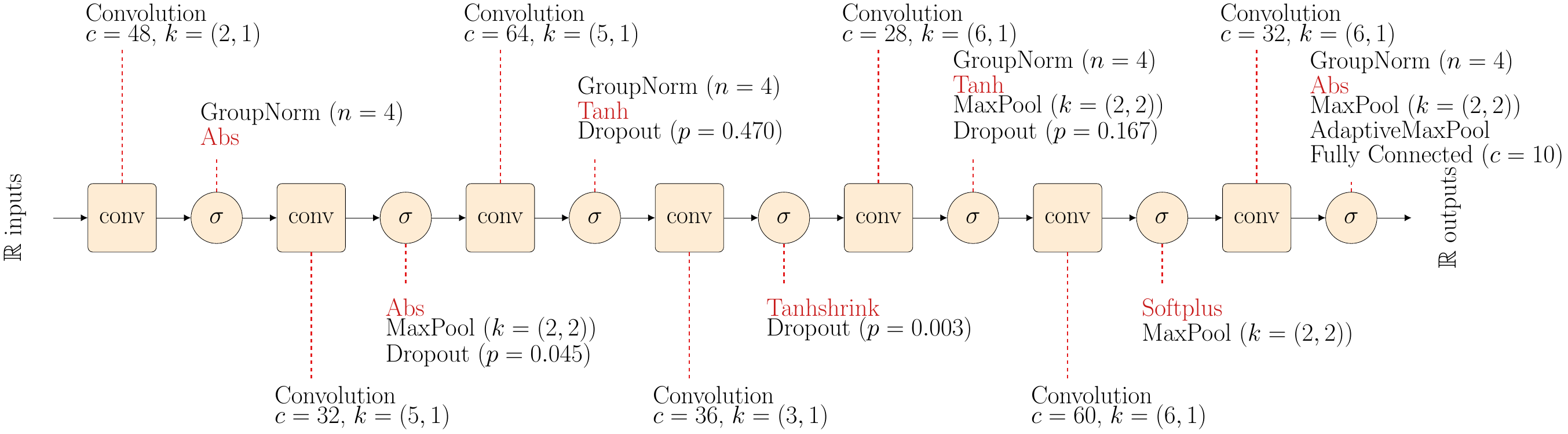}
    \vspace{2mm}
    \caption{Optimised RVNN model. $c$ is output channels, $k$ is kernel, $n$ is number of groups and $p$ is dropout rate.}
    \label{fig:real-valued-nn}
\end{figure}

\subsection{Data}

The raw audio data from AudioMNIST at 48k samples/s rate is normalised (so it's root-mean-square value equals 1) and zero-padded to 1 second of duration, where the audio content is time-shifted with a random offset for each access to satisfy time invariance requirements. White Gaussian noise is then added to the padded audio signals, with a chosen signal-to-noise ratio (SNR). An STFT of 960 points and 50\% overlap using a Hanning window is applied to the signal providing a time and frequency grid resolution of 10 ms and 50 Hz, respectively. As for the RVNN, the real and imaginary parts of the STFT are interleaved in the frequency dimension, resulting in a real-valued input. Magnitude compression of 0.5 was applied to the input data to reduce dynamic range without altering the phase. 10\% of the dataset is randomly separated for a validation set, and another 10\% for a testing set. For each audio file at the input, the output label is a class from 0 to 9. For the HNN, the data processing procedure remains largely consistent, with one notable exception: instead of interleaving the real and imaginary components of the STFT output along the frequency dimension, we utilise the complex-valued data directly as input for the HNN model.

\subsection{NAS}

As a baseline, we consider the RVNN model from~\cite{tsouvalas2022federated}, simplified to have convolutions operating in the frequency dimension of the input data. For a fair comparison, we applied a similar optimisation procedure as described in Section~\ref{sec:architecture_search}. We define, for the RVNN, three phases of optimisation: i. Customisation and block number -- the optimal configuration within each block and the number of blocks is selected, where a block in the real-valued model is composed by a convolution, an activation function and optional processing functions (pooling, normalisation and dropout); ii. Activation choice -- the activation functions for each block are selected; iii. Hyperparameters \& optimiser choice -- after finalising the structure of the RVNN, the hyperparameters, such as the learning rate, are optimised, as well as the optimiser is chosen to facilitate/enhance convergence. The considered activation functions in the real-valued architecture search procedure were: ReLU, Softplus, Tanh, Abs and Tanhshrink. 

In~\cite{sipper2021neural}, the authors employed 48 activation functions and ranked their performance, resulting in an extensive search space. To reduce the optimisation cost, we categorised these activation functions into five groups based on their shapes and selected the top-performing functions from each category in our experiments.

Additionally, to optimise the balance between memory usage and performance, in the structure refinement phase we employ a single-objective search methodology within Optuna, regarding the minimisation of the validation loss, but set constraints on the minimum and maximum number of parameters for potential architectures. This approach enables the exclusion of models which either under-perform or have an excessive number of parameters. 

The architecture of the HNN is determined as described in Section~\ref{sec:architecture_search}. For the optional function of normalisation, batch amplitude mean normalisation (BAMN)~\cite{hayakawa2018applying} is considered for complex-valued paths. The real-valued activation functions in the HNN are consistent with those employed in the RVNN, while the domain conversion and complex-valued activation functions are introduced in Sections~\ref{sec:domain_conversion} and \ref{sec:complex_activation}, respectively. 

To conserve space, we present only the architectural diagram and analysis processes corresponding to the -5 dB noise level, which represents the most challenging scenario considered, and is shown in Figure~\ref{fig:real-valued-nn} for the RVNN and Figure~\ref{fig:Hybrid-valued-nn} for the HNN.

\subsection{Performance}

Table~\ref{tab:noise_results} presents the test loss and number of parameters for the RVNN and HNN, evaluated with no noise, 0 dB and -5 dB SNR noise. For the noiseless case, the HNN could perform slightly better, in terms of test loss, than the RVNN, with a reduction of parameters of 54\%. Notably, in the -5 dB SNR scenario, where the noise level exceeds the signal strength, the number of parameters in the RVNN increases significantly. Despite this, the HNN is able to achieve much better performance with fewer parameters when compared to the RVNN. Furthermore, the performance of the RVNN degrades more rapidly with an increase in noise level, indicating HNN robustness in the presence of noise. This finding is consistent with results observed in other domains, such as carrier frequency offset (CFO) estimation, where the advantages of complex-valued operations become apparent primarily under conditions of sufficiently low SNR. 

\begin{table}[!t]
    \centering
    \caption{Comparison of RVNN and HNN performance under different noise conditions.}
    \vspace{1mm}
    \begin{tabular}{c c c c}
        \hline
        \textbf{SNR} & \textbf{Model} & \textbf{Test loss} & \textbf{Parameters} \\ \hline
        \multirow{2}{*}{\textbf{No noise}} & RVNN & 0.0066 & 39 k \\
         & HNN & $\boldsymbol{0.0052}$ & $\boldsymbol{18\ k}$ \\ \hline
        \multirow{2}{*}{\textbf{0 dB}} & RVNN & 0.0329 & 41 k \\ 
         &  HNN & $\boldsymbol{0.0097}$ & $\boldsymbol{34\ k}$ \\ \hline
        \multirow{2}{*}{\textbf{-5 dB}} & RVNN & 0.1150 & 53 k \\ 
         &  HNN & $\boldsymbol{0.0393}$ & $\boldsymbol{38\ k}$ \\ \hline
    \end{tabular}
    \label{tab:noise_results}
\end{table}

\begin{figure}[!t]
    \centering
    \includegraphics[width=0.5\linewidth]{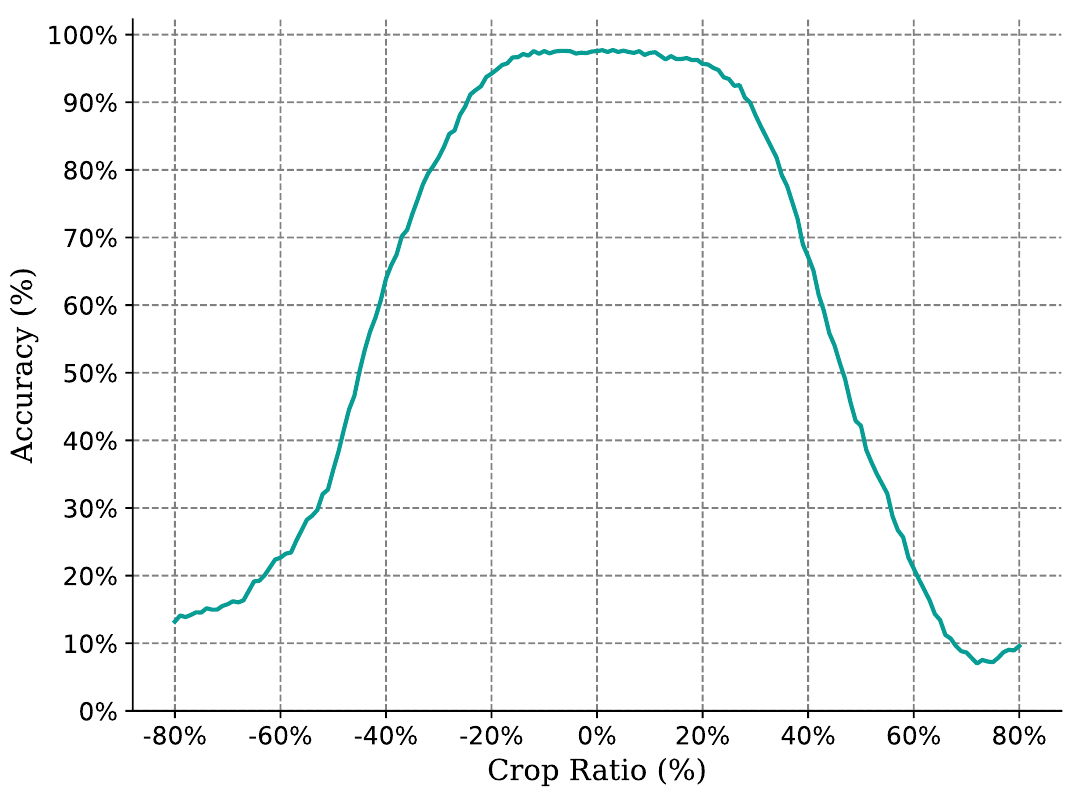 }
    \caption{Time truncation experiment for HNN from -80\% to 80\%, without added noise.}
    \label{fig:crop}
\end{figure}

To evaluate the robustness of the HNNs against incomplete input data, we conducted a time truncation experiment to test its performance under varying levels of cropping -- removing the start or the end of the audio sample. Instead of randomly padding the audio data to the length of one second, we fixed the audio data to either the start or the end part of the one-second window and shifted the data out of range by different percentages. This approach allowed us to assess how performance degrades for HNNs when a different amount of data is retained. The results are illustrated in Figure~\ref{fig:crop}, the negative crop ratio indicates the audio signal was cropped from the beginning, while the positive crop ratio indicates it was cropped from the end. The figure reveals that when the cropped ratio is less than 20\%, the performance of the HNN remains largely unaffected (probably because the margin between the start of the audio sample and the spoken digit onset). When the crop ratio is negative, the rate of performance degradation is relatively slow, indicating that the audio information at the start is more critical for the HNN in accomplishing the spoken digit classification task.

A possible explanation for the performance improvement obtained by using HNNs is that they effectively integrate real-valued and complex-valued processing paths, leveraging the advantages inherent in both domains. The complex-valued paths are capable of capturing more complicated phase relationships within the data, which can be highly beneficial for tasks involving signal processing or time series analysis. Due to the integration of complex-valued paths and the utilisation of domain conversion functions, the network may achieve more efficient parameter representation. This efficiency is particularly crucial in scenarios with limited computational resources or when attempting to mitigate overfitting. The freedom of using various layers, normalisation techniques, and activation functions provides a high degree of architectural flexibility. The network is designed through an extensive and systematic NAS procedure, as proposed in Section~\ref{sec:architecture_search}, to incorporate both real- and complex-valued processing paths, making it both robust and efficient, offering significant advantages in areas where traditional real-valued neural networks may be insufficient.

\subsubsection{Comparison to other approaches}

For the sake of comparison, we show the classification accuracy and total number of parameters for NN-based methods developed for the AudioMNIST classification task (without noise) in Table~\ref{tab:audiomnist_results}. For the proposed hybrid architecture (HNN), the number of parameters is drastically reduced when compared to the baselines. That is due to the tailored NAS procedure applied to the hybrid model, allied to the inclusion of complex operations. The accuracy obtained with the HNN remains very high even though the number of parameters is very low, showcasing the gains that can be achieved when a hybrid approach with both real- and complex-valued layers is employed.

\begin{table}[!t]
    \centering
    \caption{AudioMNIST classification (without noise) and number of parameters for the proposed and optimised HNN and other baseline approaches. The baseline values were taken from each reference.}
    \vspace{1mm}
    \begin{tabular}{lccc}
        \hline
        \textbf{Method} & \textbf{Accuracy (\%)} & \textbf{Parameters} \\
        \hline
        HNN (no noise, ours) & 98 & $\boldsymbol{18\ k}$ \\
        CNN \cite{adhishthite2023soundmnist} & 98 & 500 k to 1 M \\
        CNN \cite{bacher2023audio} & 97 & $\sim$1 M \\
        CNN \cite{cgawande2023audio} & 97 & $\sim$800 k \\
        AlexNet \cite{becker2023audiomnist} & 96 & $\sim$2 M \\
        AudioNet \cite{becker2023audiomnist} & 93 & $\sim$500 k \\
        \hline
    \end{tabular}
    \label{tab:audiomnist_results}
\end{table}

\subsection{Activation function interpretation}

As an additional analysis, we aim to interpret the choice of activation functions from the NAS process, first looking at the RVNN model. Table~\ref{tab:activations} shows the top 5 (best performing at the top, ranked by performance, 1-7 represent the building block number) activation combinations for a seven-block RVNN model. From this table we can find that in the first two blocks, the Abs activation function is overwhelmingly selected in the top three combinations. This suggests that the absolute value operation might be particularly effective in the initial stages of the network. The Abs function likely aids in extracting the amplitudes of the complex-valued input, which might simplify the learning process in the subsequent layers. In the later modules ``Tanh'' (saturation) and ``Tanhshrink'' (thresholding) activation functions are consistently used, which can be seen as complementary functions. This combination may offer a balanced approach to managing signal amplitude while maintaining non-linearity, helping in the refinement of feature maps. While ReLU is well-known for its simplicity and efficiency, its selective use suggests that it may only be effective when combined with other functions like Abs and Tanh in this specific case. Softplus, a smooth approximation of ReLU, appears to be preferred in deeper layers, possibly due to its ability to support stable learning and smoother gradient propagation.
In the HNN case, interpretation can be aided by observing the separation of information into distinct pathways and examining the function choices within these partial networks as above. Interestingly, the formation of similar pathways over multiple NAS sessions indicates a recurring solution pattern.

\begin{table}[!t]
    \centering
    \caption{Top 5 activation function combinations selected per block by Optuna for the RVNN at -5 dB SNR.}
    \vspace{1mm}
    \begin{tabular}{c c c c c c c}
        \hline
        \textbf{1} & \textbf{2} & \textbf{3} & \textbf{4} & \textbf{5} & \textbf{6} & \textbf{7} \\
        \hline
        Abs   & Abs       & Tanh  & Tanhshrink & Tanh  & Softplus & Abs \\
        Abs   & Abs       & Tanh  & Tanhshrink & Tanh  & Softplus & ReLU \\
        Abs   & Abs       & Tanh  & Tanhshrink & Tanh  & Softplus & Abs \\
        ReLU  & Tanh      & ReLU  & Tanhshrink & ReLU  & Abs      & Abs \\
        ReLU  & Tanh      & Tanh  & Tanhshrink & Abs   & Softplus & Tanhshrink \\
        \hline
    \end{tabular}
    \label{tab:activations}
\end{table}

\section{Discussion}
\label{sec:discussion}
We have described one method of architecture search, however, we see alternative approaches that could be used, e.g., starting with a single real-valued path, a HNN could ``grow" complex-valued pathways to expand the network and become more efficient in terms of loss and parameter storage. This represents a practical option for expanding existing real-valued networks to their hybrid counterpart.

Given an optimised architecture, with specific routing, selected activation and domain conversion functions, this could indicate novel solution spaces not so easily identified within real-valued networks due to their homogeneous nature. The separation of real- and complex-valued information pathways could aid in explainability. Using targeted stimuli injected into the network, pathways and functions could be identified more easily in decoding how the network solved the problem under study.

The generalised activation functions \eqref{eq:activation_P} and \eqref{eq:activation_Ps} offer the expressivity of polynomials (when $|z|^q < \epsilon$) and self limiting from the denominator term (when $|z|^q > \epsilon$), this is an alternative to the use of splines.
On one hand, keeping the coefficients real-valued, we can generalise real-valued activation functions as bounded polynomials and on the other hand, using complex valued coefficients, generalisation into the complex space is facilitated. Inspired by Kolmogorov-Arnold network (KAN) architectures \cite{liu2024kankolmogorovarnoldnetworks}, HNNs can offer a middle ground between KAN, real- and complex-valued architectures by allocating coefficients between the linear (convolutions) and non-linear (activation) functions.

Expanding this work to other models like variational autoencoders~\cite{pinheiro2021variational}, a potential approach is to modify the encoder and decoder into hybrid real- and complex-valued counterparts, using a block similar to Figure~\ref{fig:hybrid_block}. The latent space could include real-valued variances and real- and complex-valued means at the output of the encoder. In a similar way, for a transformer structure -- initially proposed by \cite{vaswani2017attention} -- processing blocks like convolutional and fully connected layers could be expanded into a hybrid version as in Figure~\ref{fig:hybrid_block}, while the multi-head attention mechanism block could be further developed into a hybrid version with information exchange between paths. Notice that a complex-valued version of the multi-head attention mechanism was already developed \cite{Tokala2024timefrequency}, which can serve as a baseline together with its real-valued version. In both cases, it would be naturally interesting to apply a NAS procedure such as the one proposed in Section~\ref{sec:architecture_search}, drastically reducing the amount of parameters and improving the model's performance.

\section{Conclusion}
\label{sec:conclusion}
We have analysed how an RVNN processed complex-valued data and, with these insights, we have proposed a HNN architecture and a tailored neural architecture search procedure. Comparing an RVNN and a HNN applied to a practical problem, we have demonstrated a significant performance improvement.
%
The HNN deviates from the usual real domain solution space and importantly provides for native use of complex- and real-valued data.
The novel complex-valued activation and domain conversion functions offer great flexibility in combination with the optimisation framework to explore the HNN architecture space. The obtained results show a great promise in terms of increased performance and model size reduction with the hybrid architecture, which successfully leverages real- and complex-valued processing. In comparison to existing methods, the HNN can achieve state-of-the-art classification accuracy for the problem of spoken digits recognition with a dramatically lower number of parameters, due to the leveraging of complex operations allied to a dedicated NAS framework.

The hybrid architecture is proposed relying on an extensive and systematic neural architecture search procedure to reduce its size and optimise its performance. As a future line of research in architecture development, a simplification of the model design can be investigated. Regarding the HNN application, future works could consider other problems of a complex-valued nature such as audio/speech enhancement or synthesis, beamforming, biomedical-related analysis based on electrocardiogram and electroencephalogram signals, etc.








\section*{Declarations}

\subsection*{Acknowledgments and funding}
This work was supported by the Robust AI for SafE (radar) signal processing (RAISE) collaboration framework between Eindhoven University of Technology and NXP Semiconductors, including a Privaat-Publieke Samenwerkingen-toeslag (PPS) supplement from the Dutch Ministry of Economic Affairs and Climate Policy.

\subsection*{Author's contribution}
Alex Young conceptualised the study, designed the methodology, interpreted results, wrote original software, conducted experiments and wrote the original draft. Luan Vinícius Fiorio designed the methodology, wrote original software, conducted experiments and wrote the original draft. Bo Yang performed software development, execution of experiments, analysis of results and wrote the original draft. Boris Karanov provided critical feedback and revision. Wim van Houtum provided supervision, critical feedback and project administration. Ronald M. Aarts provided supervision and critical feedback.

\subsection*{Data availability}
The data necessary for reproduction of results is publicly available at \url{https://github.com/soerenab/AudioMNIST} \cite{becker2023audiomnist}. The proposed methods and experiments can be completely reproduced based on the text, equations, figures and tables in the manuscript. The tools used to obtain results, Python programming language (\url{https://www.python.org}) and PyTorch (\url{https://pytorch.org}) \cite{paszke2019pytorchimperativestylehighperformance}, are free to use and publicly available.

\subsection*{Competing interests}
The authors declare that they have no competing interests.

\bibliographystyle{IEEEtran}
\bibliography{template}

\end{document}